# A Trajectory-free Crash Detection Framework with Generative Approach and Segment Map Diffusion

Weiying Shen, Hao Yu, Yu Dong, Pan Liu*, Yu Han, Xin Wen
School of Transportation, Southeast University, Nanjiang, China
{shenweiying, seuyu, dongyu_007, liupan, yuhan, wenxin}@seu.edu.cn
* Corresponding author

## ABSTRACT

Real-time crash detection is essential for developing proactive safety management strategy and enhancing overall traffic efficiency. To address the limitations associated with trajectory acquisition and vehicle tracking, road segment maps recording the individual-level traffic dynamic data were directly served in crash detection. A novel two-stage trajectory-free crash detection framework, was present to generate the rational future road segment map and identify crashes. The first-stage diffusion-based segment map generation model, Mapfusion, conducts a noisy-to-normal process that progressively adds noise to the road segment map until the map is corrupted to pure Gaussian noise. The denoising process is guided by sequential embedding components capturing the temporal dynamics of segment map sequences. Furthermore, the generation model is designed to incorporate background context through ControlNet to enhance generation control. Crash detection is achieved by comparing the monitored segment map with the generations from diffusion model in second stage. Trained on non-crash vehicle motion data, Mapfusion successfully generates realistic road segment evolution maps based on learned motion patterns and remains robust across different sampling intervals. Experiments on real-world crashes indicate the effectiveness of the proposed two-stage method in accurately detecting crashes.

## 1. Introduction

Crash detection is a critical component in traffic management systems. Timely and accurate crash identification enables rapid emergency response, potentially mitigating injury severity and property damage. Furthermore, efficient detection facilitates proactive traffic control measures upstream of the crashes, minimizing network-wide disruptions.

The traffic crash detection methods evolved alongside the sensor technologies. Traditionally, the crash detection methods relied on fixed-location microwave detectors collecting macroscopic traffic parameters, e.g., volume, speed. For instance, the classic California algorithm (*1*), one of the most widely spread rule-based approaches, identifies crashes through abnormal spatiotemporal fluctuations in these parameters. Subsequent refinements employed statistical classifiers (*2*, *3*) and the machine learning (ML) classifiers (*4*–*6*) to improve the accuracy. However, crash detection based on macroscopic traffic parameters has a major limitation, i.e., the inability to accurately locate the involved crash vehicles (*7*). With the popularity of closed-circuit televisions, the video-based crash detection approaches have been widely developed. These approaches include classical image processing techniques, such as optical flow models and Gaussian mixture models (*8*, *9*), and deep learning methods designed to extract crash features(*10*, *11*). While videos provide a more intuitive localization than macroscopic traffic data, they still struggle to distinguish congestion from minor crashes due to overlapping fields of view and perspective constraints (*12*).

With the development of high-frequency radar and computer vision technology, vast amounts of individual-level traffic dynamic data have been accumulated, supporting faster crash detection and more precise localization (*13*). These traffic dynamic data include both micro-level traffic information obtained from radar sensors and high-resolution videos captured by surveillance cameras. Crash detection based on individual-level traffic dynamic data can be briefly classified into two main categories: single-vehicle based methods (*14*, *15*) and multi-vehicle based methods (*7*, *16–18*). The single-vehicle based methods typically assess crashes via threshold comparisons of single-vehicle metrics, such as the speed, the distance between adjacent vehicles, and the maximum deceleration. However, due to the inherent heterogeneity of driver behavior and the high uncertainty in the correlation between the detected traffic data and actual traffic conditions, the reliability and completeness of these approaches remain questionable (*16*). In cope with this issue, the multi-vehicle based approaches were benefit for constructing the traffic conditions among adjacent vehicles, which are mainly the ML methods and deep learning methods. In these methods, crash detection is formulated as a binary classification issue, while the input information involves the vehicle motion before and after a crash (*7*, *18*).

Noted that most existing crash detection methods relies on continuous vehicle motions information, typically the trajectory data. However, constructing smooth and reliable vehicle trajectories from individual-level traffic data is challenging. For instance, the tracking process is easily influenced by noise and fluctuations in the data collecting process (*19*), leading to a sudden change of the vehicle IDs. As a result, trajectories of one vehicle are recorded as discontinuous records of several vehicles. Moreover, recent multi-vehicle based methods usually focus on only a certain number of vehicles which are directly involved during car-following or lane-changing scenarios. These approaches, however, may overlook the latent interconnected information embedded in the collective movement of vehicle trajectories over continuous motion dynamics. Consequently, a substantial amount of information that could potentially improve the accuracy of crash detection is lost.

In essence, crash risk is a combination of the spatiotemporal relationships of all the traffic participants. It originates from an unobservable behavioral coordination among these traffic participants, and finally leads to failures in space-time trajectory intersections. In order to track the crash risk propagation and to enhance the accuracy of crash detection, it is of great importance to capture the spatiotemporal evolution of the traffic participants. Accordingly, the authors proposed the crash detection framework based on a novel designed vehicle motion information structure, termed as road segment map (RSM). A RSM is an image-like data structure, which records the spatial relationship for all the traffic participants within a road segment at each time slice. Here, the RSM is not limited to a specific type of vehicle motion information. It can encompass video data, point cloud data, radar echo data, depth map, trajectory data, or simply the location data. Take the location data for example, as shown in Figure 1(a). The source information comes from the video that record vehicle motion. After identifying the position of each traffic participant, each frame of the video forms a RSM. Similarly, as shown in Figure 1(b), road segment maps can be constructed from multi-source data. By preserving the spatiotemporal characteristics of each vehicle through sequential image arrangement, road segment maps enable crash detection without requiring inter-frame vehicle matching and tracking. This representation provides a higher-level perspective for crash detection and adapts well to low-frequency sensor data.

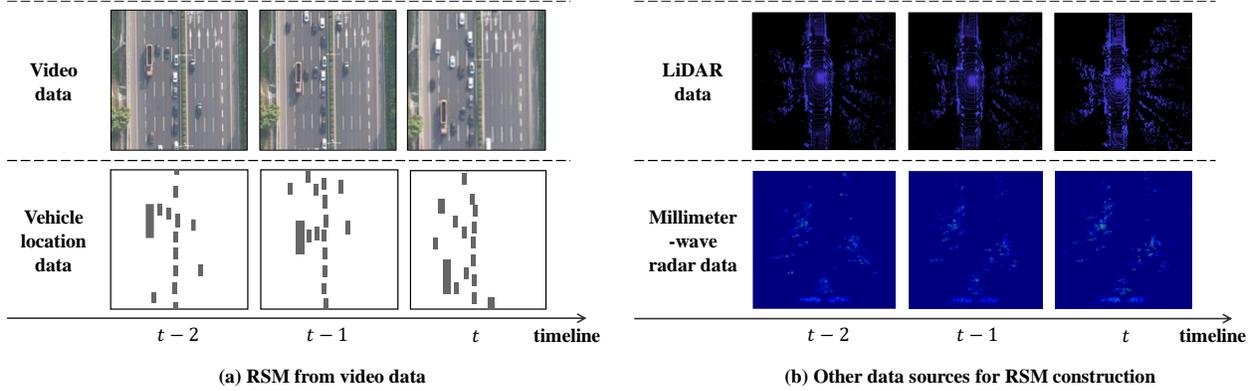

**Figure 1.** Road segment maps from diverse spatiotemporal data sources.

Leveraging the spatiotemporal information embedded in road segment maps, a two-stage crash detection framework is proposed in this study, including a RSM-evolution part and a crash detection part. More specifically, a diffusion-based model onto the road segment maps (Mapfusion) is firstly trained on naturalistic driving dataset to predict the maps of next time slice based on historical data, considering the RSM at any given moment can be significantly diverse due to the inherent heterogeneity of drivers and the variability of driving behaviors (*20*, *21*). In the second stage, a discrepancy analysis is conducted between the predicted crash-free maps and the observed one to identify crashes.

The contributions of this study can be summarized as follows: Firstly, road segment maps were utilized as input for crash detection algorithm, retaining information of multiple traffic participants from raw traffic data. This approach mitigates the challenges of acquiring uninterrupted vehicle trajectory data while preserving individual-level traffic information for crash detection. Secondly, a data-driven approach crash detection was introduced based on the generation of non-crash scenarios. The generative model leverages naturalistic non-crash datasets to train its ability to identify crashes, with less requirement on crash-specific datasets. Thirdly, the proposed diffusion-based architecture effectively combines diffusion models with sequential embedding components to introduce the spatiotemporal features and improve the plausibility of vehicle data in the generated road segment maps.

## 2. Problem description

A set of road segment maps, $\{R^n\}$, are used to detect vehicle crashes in real-time using individual-level traffic dynamic data from a targeted area. For clarity, Table 1 presents the variables used throughout this paper.

For a targeted area, the motion variables of all vehicles within the road segment, such as position, speed, and heading angle, are captured by detectors installed along the road. These traffic dynamics are encoded into road segment maps, represented as $R^n \in \mathbb{R}^{C \times H \times W}$, where *n* denotes the time of the segment map. Let $\mathcal{R} = \{R^t\}_{t=n-f}^{n-1} \in \mathbb{R}^{f \times C \times H \times W}$ denote the observed sequence of road segment maps within a time window of length *f*, which serves as the input to our crash detection framework. Suppose *M* represents the static background information of the monitored road segment, including lane geometries, directions, and other travel-related attributes. Based on the input historical frames $\mathcal{R}$ and the background $M$, the framework generates the predicted road segment map $\hat{R}^n$ for the next time $n$. Crash detection is to identify the time when the observed segment map exhibits significant deviations from the non-crash generations $S_{\hat{R}^n}$. This comparison is conducted in the feature space, by measuring discrepancies between the feature representations $R_F$ and $\hat{R}_F$.

**Table 1.** Notations and descriptions.

| Notation | Description |
|---|---|
| $R^n$ | Real road segment map for the detection area at timestep $n$ |
| $\hat{R}^n$ | Generated road segment map for the detection area at timestep $n$ |
| $f$ | Length of the detection time window |
| $\mathcal{R}$ | Sequence of known road segment maps in the detection time window, $\mathcal{R} = \{R^t\}_{t=n-f}^{n-1}$ |
| $S_{\hat{R}^n}$ | Road segment maps sample set from generation model at timestep $n$ |
| $R_F$ | Features extracted from real road segment map |
| $\hat{R}_F$ | Features extracted from generation road segment map |
| $M$ | Static background of the detection road segment |

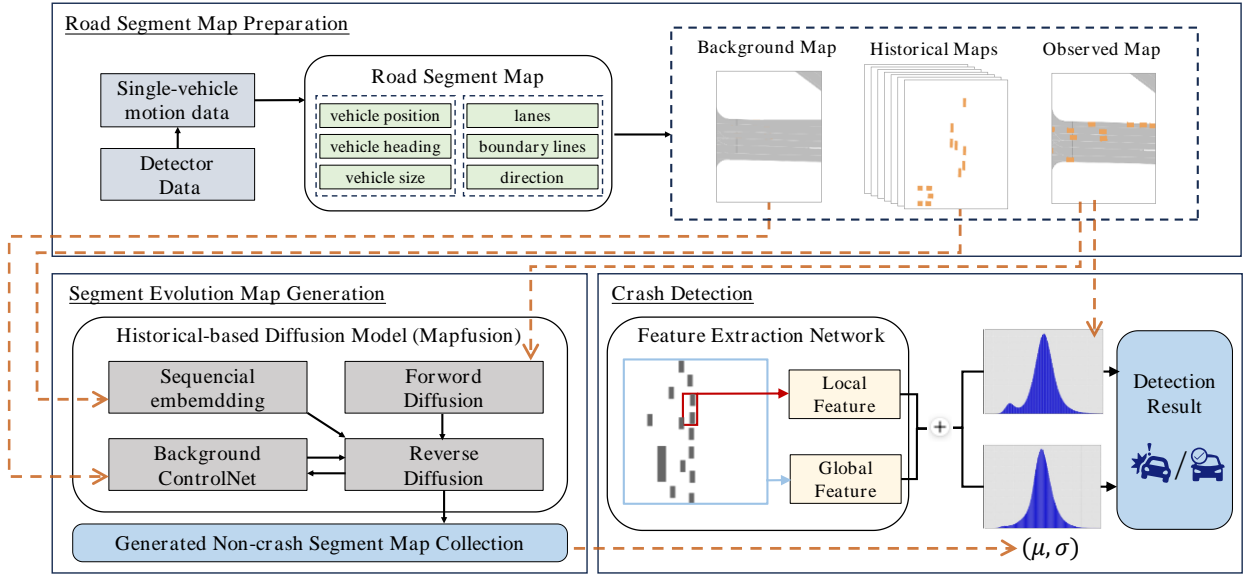

**Figure 2.** The overall framework.

## 3. Methodology

As illustrated in Figure 2, the proposed crash detection framework consists of three parts, including road segment map preparation, segment map evolution generation and crash detection. Road segment map preparation transforms raw sensor data into a standardized road segment map and background map, which serve as the inputs for subsequent modules. The segment maps are divided into two parts: a sequence of historical maps over a fixed time window, and the current observed map. The background map and the sequence of historical segment maps are leveraged to establish Mapfusion. Mapfusion generates a set of segment maps that represent non-crash scenarios at the current time, reflecting rational vehicle evolution patterns. Finally, crashes are identified by measuring feature differences between the observed map and generated non-crash segment maps.

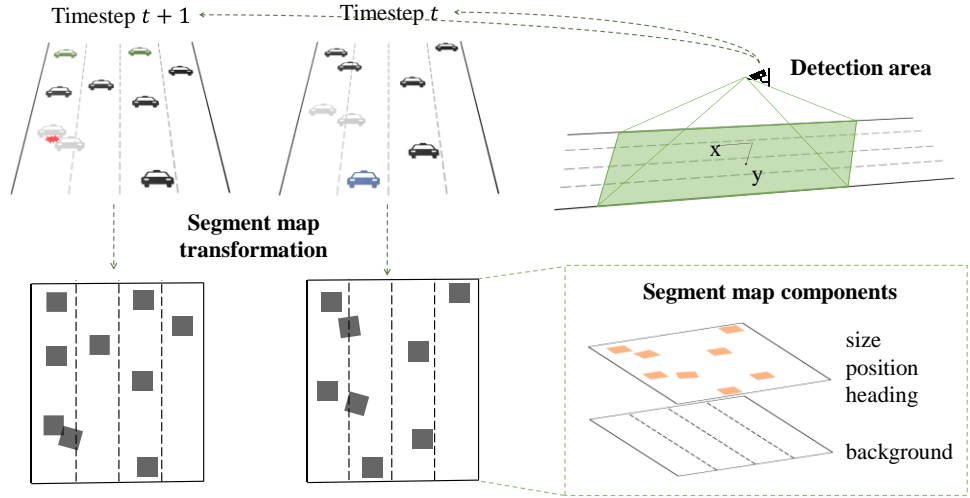

**Figure 3.** The road segment maps code process.

### 3.1. Road segment map preparation

To enable unified model input across different sensors, we propose a standardized method for constructing road segment maps based on vehicle motion data. Each map encodes the position, size, and heading of every vehicle on the road at a specific time. The raw numerical data are transformed into a structured image representation, where each vehicle is encoded at its corresponding pixel location. Importantly, the segment maps focus exclusively on vehicles on the detected road segment, omitting irrelevant environmental elements outside the road boundaries and non-vehicle road users.

Figure 3 demonstrates the process of encoding road segment maps into RGB images. The spatial boundary of each road segment map is determined by the sensing coverage of the deployed sensors. For each vehicle detected at a given moment, its position is mapped to a corresponding pixel coordinate in the image by proportionally scaling its distance relative to the predefined map boundaries. A rectangular representation of the vehicle is then rendered at this coordinate, with its dimensions corresponding to the length and width of the vehicle. The headings of the vehicles are reflected by rotating the rectangle according to the recorded angle. Finally, a background layer representing lane markings and road structure is fused beneath the vehicle layer. Alternatively, the vehicle and background information can be encoded into two separate images to accommodate varying input requirements across different model architectures.

Speed information is not directly embedded in the segment maps, for two primary reasons: (1) Video-based sensors such as surveillance cameras and unmanned aerial vehicles typically require additional post-processing to estimate vehicle speed. As a result, including speed may introduce additional errors due to data noise or imprecise processing algorithms. (2) As the vehicle positions are sequentially encoded in temporally ordered road segment maps, velocity can be effectively inferred from inter-frame displacements. Spatiotemporal architectures are capable of capturing motion dynamics by learning patterns of positional transitions over time, even in the absence of velocity.

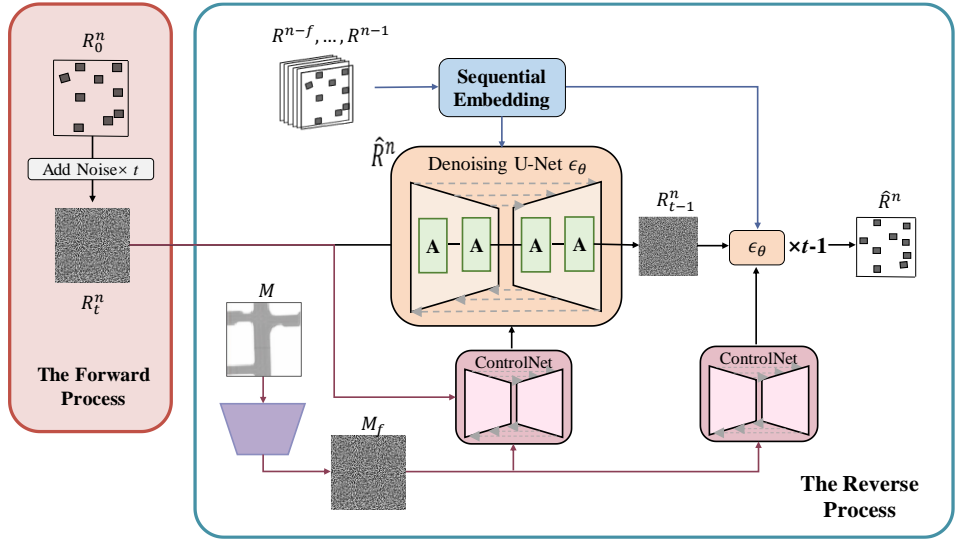

**Figure 4.** The structure of Mapfusion.

### 3.2. Segment evolution map generation

The wide range of possible road segment map at any given moment is difficult for existing deterministic crash prediction models to fully capture. Therefore, it becomes necessary to adopt a generative modeling approach. The generative models can learn the underlying probability distribution of input data and create new data similar to real-world observations (*22–24*). Such models enable the generation of diverse vehicle states on the road segment.

Diffusion models, characterized by their superior sample quality, diversity, and controlled image generation capabilities, offer a compelling alternative in the field of image generation (*25*). However, the application of diffusion models in crash detection remains relatively unexplored. The ability of diffusion models to generate images with realistic distributions applies to the RSM and holds potential for enhancing the efficacy of crash detection. This potential can be further realized by incorporating prior knowledge to guide the generation process. Previous research has achieved great success in using diffusion models to generate conforming images based on prior knowledge of text prompt, such as Stable Diffusion (*26*) and Dalle2 (*27*). The Stable Diffusion model integrates text with images in the cross-attention components during training, improving the alignment between the generated image and the text. Additionally, Video Diffusion Model from Google (*28*) enables the generation of 3D videos by incorporating temporal information into the model training process, demonstrating that temporal features can be integrated into both the training and sampling processes of diffusion models.

Therefore, in this section, we use Mapfusion to generate road segment maps that align with realistic evolutionary distributions. It consists of a forward process and a reverse process. Specifically, the forward process of the diffusion model gradually adds noise to the actual segment map to corrupt its distribution until it transitions to pure Gaussian noise. The reverse process then restores the original road segment map from this fully degraded state. The detailed network architecture of Mapfusion is illustrated in Figure 4. As described in the problem statement, generating road segment maps requires incorporating two distinct types of information to enhance the realism of the generated outputs: sequential road segment maps and a background map. In our model, we employed two different strategies to integrate these information sources effectively.

Mapfusion has two main components: a denoising U-Net and a ControlNet. ControlNet is a neural network designed to regulate a pre-trained image generation model by incorporating additional conditioning information, thereby enabling more controlled and realistic image generation within the diffusion model(29). In this study, the background image serves as the conditioning input. Before ControlNet can be effectively integrated, a foundational diffusion model must first be trained.

The training process primarily focuses on optimizing the denoising U-Net, which serves as the backbone of the diffusion model. Notably, the forward process does not require learning any parameters, as it merely involves noise perturbation. It incrementally adds noise $\mathcal{N}(0, \mathbf{I})$ to the original data distribution. In our model, noise is added to the segment map at the detection timestep $n$ as $q(R_0^n)$. The forward process satisfies the Markovian structure with the latent variables $\{R_t^n\}_{t=1}^T$. Therefore, the conditional probability distribution of the noisy segment map $R_t^n$ at time $t$ is defined as follows,

$$q(R_t^n|R_0^n) = \mathcal{N}(R_t^n; \sqrt{\bar{\alpha}_t} R_0^n, (1-\bar{\alpha}_t)\mathbf{I}) \tag{1}$$

where $\alpha_t = 1 - \beta_t$, $\bar{\alpha}_t = \prod_{i=1}^t \alpha_t$, and $\beta_t$ is hyperparameter of the variance of the Gaussian distribution.

The reverse process is to learning to denoise $R_t^n \sim q(R_t^n|R_0^n)$ as $p_\theta(R_{t-1}^n|R_t^n) = \mathcal{N}(R_{t-1}^n; \mu_\theta, \sigma_\theta^2)$ for all $t$, where $\theta$ represent the updated parameters of the denoising U-Net model $\epsilon_\theta$, resulting in $\mu_\theta = \frac{1}{\sqrt{\alpha_t}}(R_t^n - \frac{\beta_t}{\sqrt{1-\bar{\alpha}_t}}\epsilon_\theta)$. Since our model aims to generate the segment map at timestep $n$ based on its evolution, the sequence of segment maps for the previous $f$-frames, $\mathcal{R}^{n-f:n-1} = \{R^{n-f}, \dots, R^{n-1}\}$, need to be inputted into the denoising model as prior knowledge to help the network improve the reliability of the generation. The modification made in the diffusion model with the conditional generation is to provide $R^{n-f:n-1}$ to the model as $p_\theta(R_{t-1}^n|R_t^n, \mathcal{R}^{n-f:n-1})$. Therefore, our goal of the training phase is to minimize the variational upper bound of the negative log-likelihood as follows:

$$\mathbb{E}_{R_t^n, t, \mathcal{R}^{n-f:n-1}, \epsilon}[\|\epsilon - \epsilon_\theta(R_t^n, t, \mathcal{R}^{n-f:n-1})\|^2] \tag{2}$$

Based on the trained denoising U-Net, ControlNet directly inherits all parameters from the pre-trained denoising U-Net and takes the processed RGB background map images as its input for further training. During this training phase, the parameters of the original denoising U-Net remain entirely fixed, while only the parameters within the ControlNet are progressively optimized. This approach ensures that the original model's generative capabilities remain unaffected while allowing for enhanced control over the generated output, aligning it more closely with the input conditions. The forward process still follows Equation 1, while the ControlNet structure is applied at each step of the reverse process alongside the trained U-Net. Consequently, a new conditioning term is introduced into the loss function, which is modified as follows:

$$\mathcal{L} = \mathbb{E}_{R_t^n, t, \mathcal{R}^{n-f:n-1}, M, \epsilon}[\|\epsilon - \epsilon_\theta(R_t^n, t, \mathcal{R}^{n-f:n-1}, M)\|^2] \tag{3}$$

After the final training of diffusion model, the reverse process can be directly applied. It employs the resampling technique step by step from $R_T^n \sim \mathcal{N}(0, \mathbf{I})$ to generate the realistic evolved segment map at timestep $n$. This process is called sampling. In the sampling process, the sequence of the segment maps for the previous $f$-frames is still required to be condition, defined as $p_\theta(R_{t-1}^n|R_t^n, \mathcal{R}^{n-f:n-1}, M) = \mathcal{N}(R_{t-1}^n; \mu_\theta(R_t^n, t, \mathcal{R}^{n-f:n-1}, M), \sigma_\theta^2)$. Based on the above model structure, the detailed procedure process of training and sampling are described in Algorithm 1 and Algorithm 2.

**Algorithm 1** Mapfusion Training

1: **repeat**
2: $R_t^n \sim q(R_0^n)$
3: $t \sim \text{Uniform}(\{1, ..., T\})$
4: $\mathcal{R}^{n-f:n-1} = \{R^{n-f}, ..., R^{n-1}\}$
5: $\epsilon \sim \mathcal{N}(0, \mathbf{I})$
6: Take gradient descent step on
$$\nabla_\theta \left\| \epsilon - \epsilon_\theta(R_t^n, t, \mathcal{R}^{n-f:n-1}) \right\|^2$$
7: **until** converged
8: **repeat**
9: $R_t^n \sim q(R_0^n)$
10: $t \sim \text{Uniform}(\{1, ..., T\})$
11: $\mathcal{R}^{n-f:n-1} = \{R^{n-f}, ..., R^{n-1}\}$
12: $M$
13: $\epsilon \sim \mathcal{N}(0, \mathbf{I})$
14: Take gradient descent step on
$$\nabla_\theta \left\| \epsilon - \epsilon_\theta(R_t^n, t, \mathcal{R}^{n-f:n-1}, M) \right\|^2$$
15: **until** converged

---

**Algorithm 2** Mapfusion Sampling

1: $R_T^n \sim \mathcal{N}(0, \mathbf{I})$
2: **for all** *t* from T to 1 **do**
3: $R_{t-1}^n \sim \mathcal{N}(\mu_\theta(R_t^n, t, \mathcal{R}^{n-f:n-1}, M), \sigma_\theta^2(R_t^n, t, \mathcal{R}^{n-f:n-1}, M))$
4: **end for**
5: **return** $R_0^n$

---

To better integrate sequential road segment map information from previous frames into the diffusion model and ensure full utilization of their temporal relationships between $R_t^n$ and $\mathcal{R}^{n-f:n-1}$, a sequential embedding module was introduced to extract sequence features, as shown in Figure 5. Inspired by Shi et al. (*30*), this module employs Convolutional LSTM (ConvLSTM) cells to encode the segment map sequence. As a variant of the long short-term memory (LSTM) model, ConvLSTM achieves great performance in spatiotemporal sequence forecasting problems. The segment map sequence $\mathcal{R}^{n-f:n-1}$ serves as the input to the ConvLSTM encoder, where each state represents the temporal evolution of previous segment maps. By learning these states, the encoder effectively captures the historical evolution of road segments. To ensure the generated segment maps align with previous frames, a cross-attention mechanism is incorporated to establish dependencies between the segment map to generate and input sequences. This mechanism is integrated into the downsampling, mid-block, and upsampling of the denoising U-Net model in diffusion model. At each denoising step, the latent noise segment map receives sequence information to improve the accuracy of the generated road segment maps.

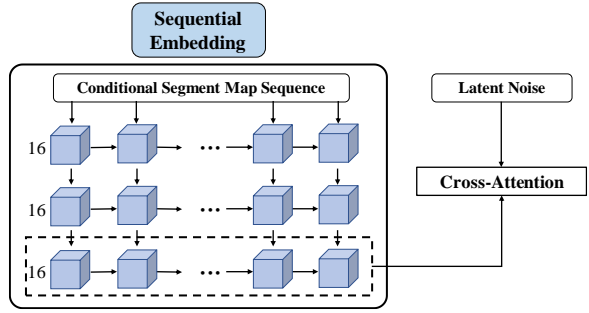

**Figure 5.** CovnLSTM Sequential Embedding.

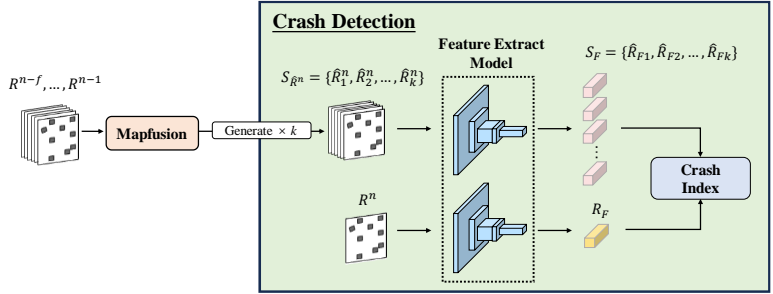

**Figure 6.** Architecture design of crash detection.

### 3.3. Crash detection

Building on the segment evolution map generation, this section presents the methodology for crash detection using the developed model in section 3.3. The overall design of the proposed framework is shown in Figure 6.

The diffusion model enables simultaneous sampling to generate multiple target samples. Therefore, within a crash detection time window, the trained diffusion model takes the previous *f* frames of road segment maps, from timestep *n-f* to *n-1*, as input. Then, it generates *k* road segment maps of timestep *n*, forming a sample population $S_{\hat{R}^n} = \{\hat{R}_1^n, \hat{R}_2^n, ..., \hat{R}_k^n\}$ that represents the expected non-crash road segment maps. It can be inferred that the feature distributions within the normal sample set should be relatively similar. To detect potential abnormalities, the generated sample population $S_{\hat{R}^n}$ is compared with the monitored road segment map $R^n$. $R^n$ is captured in real-world at timestep n. If the feature distribution of the monitored road segment map closely aligns with that of the generated normal sample set, it is assumed to belong to the normal evolution pattern. Conversely, it may indicate an abnormal situation.

To quantify the differences between road segment maps, we refer to the Fréchet Inception Distance (FID), a widely used evaluation metric in image generation. FID is a metric for measuring the discrepancy between the outputs of the generated model and the real images' distribution, proposed by Martin Heusel *et al* (*31*). However, the monitored road segment map $R^n$ is unique with a sample size of one. To address this limitation, a similar feature difference metric was created to quantify the discrepancy between the monitored road segment map feature $R_F$ and the generated sample features $S_F = \{\hat{R}_{F1}, \hat{R}_{F2}, ..., \hat{R}_{Fk}\}$. This feature difference index is defined as the *Crash Index*. Equation 4 defines the calculation of the *Crash Index* at the timestep *n*, where the first term measures the overall feature distance and the second term captures local feature variations:

$$a^n = \frac{1}{C}\|\bar{S}_F - R_F\|^2 + \frac{1}{\lambda}\sum \max_\lambda (\bar{S}_F - R_F)^2 \tag{4}$$

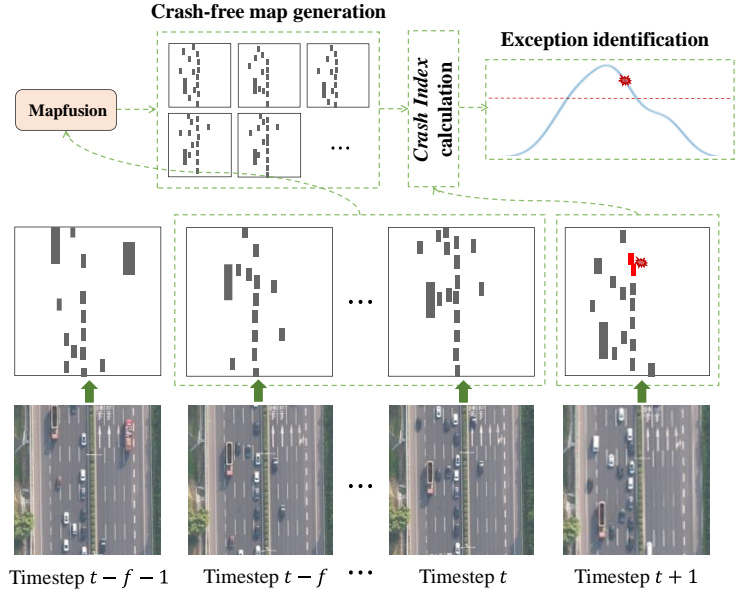

**Figure 7.** Crash detection process based on Mapfusion and *Crash Index.*

where $\bar{S}_F$ denotes the feature-wise mean of the generated segment maps; $C$ represents the number of feature channels; $\lambda$ is the number of feature dimensions selected to measure local mutations in the segment map.

When the *Crash Index* is relatively low, the feature discrepancy between the monitored road segment map and the ensemble of generated road segment maps is correspondingly minimal, indicating a lower probability of a crash at the monitored timestep $n$. To effectively identify crash occurrences, an *Crash Index* threshold $\gamma$ is established based on the distribution of road segment maps at labeled crash timesteps. A crash is detected when the *Crash Index* surpasses the threshold:

$$\mathcal{D}^n = I(a^n > \gamma) \tag{5}$$

where $I(\cdot)$ outputs 1 if the condition holds, otherwise 0. Thus, $\mathcal{D}^n \in \{0,1\}$ is the binary crash indicator. The threshold $\gamma$ can be estimated using quantile analysis. Real-time crash detection is achieved by continuously advancing the monitoring time window of the road segment map and analyzing variations in the *Crash Index*. The approach for crash detection based on *Crash Index* is shown in Figure 7.

Since the *Crash Index* quantifies the degree of feature divergence in road segment maps, the choice of an effective feature extraction network is critical. The Vision Transformer (ViT)(*32*) leverages a self-attention mechanism to capture both global and local positional relationships within an image. This capability makes it particularly appropriate for processing road segment maps, which contain rich spatial and directional information. We adopt a pretrained ViT model with its classification head removed, allowing it to output high-dimensional representations of road segment features. These features encode spatial interactions between vehicles, as well as their positional distribution within the segment. The extracted representations are subsequently used in Equation 6 to compute the final *Crash Index*.

## 4. Numerical Tests

### 4.1. Data description

Numerical experiments were designed to verify the performance of segment map generation and crash detection capabilities of the model. Experiments were conducted on two complementary datasets: a public motion forecasting dataset without crash scenarios for segment map generation evaluation, and a self-collected crash dataset for crash detection analysis. These datasets were specifically chosen to address the two phase-based objectives of our research framework. All experiments were carried out on an Ubuntu 20.04 computer equipped with an Intel(R) Core(TM) i7-12700KF CPU and a GeForce RTX 4090 GPU. Code was written under the PyTorch.

*4.1.1. Motion forecasting dataset without crash scenarios*

Although the network proposed in this paper uses individual-level traffic dynamic data as input, current open-source vehicle motion datasets already contain trajectory information such as vehicle ID. Therefore, by omitting information containing trajectories, the feasibility of the proposed model for the segment evolution map generation was tested based on the Argoverse 2(AV2) motion forecasting dataset, a dataset for perception and prediction studies of autonomous driving (*33*). The dataset contains 250,000 scenarios with 5-second trajectories and corresponding HD maps covering intersections, roundabouts, and highways. Only the entries within the motion forecasting dataset with the target category of vehicle are collected to code segment map, including variables the coordinate position, heading angle, velocity and category of road users on the scenarios. Experiments were conducted using the training, validation, and test sets partitioned from the AV2 Motion Forecasting Dataset. A reconstruction method inspired by the data visualization method provided by AV2 is applied to process the datasets and draw the road segment maps at the sampling rate of 10 Hz. 11 frames of the road segment map sequence are input to the model while the first 10 were used as a priori conditions for training and sampling, and the last frame was used as the ground truth that needs to be generated. The size of the road segment maps in the experiments are processed to 512×512.

*4.1.2. Field collected data in Hurong highway covers the entire process of traffic crash*

Individual-level traffic dynamic data before and after the traffic crashes is needed in this study to verify the reliability of the crash detection method based on the Mapfusion. However, existing public datasets lack fine-grained crash annotations. To address this gap, we collected real-world crash events captured from aerial views based on drone. Due to the scarcity of crashes in the real world, only two aerial videos before and after crashes occurring on the Hurong highway are accessible in this study to close the loop of the crash detection method. The exact timestep of the two crashes are recorded as labels for the crash detection. Previous studies on crash detection and crash warning have typically employed data ranging from 0.1 to 4 seconds. The time window between the emergence of discernible crash precursors and the actual occurrence of a crash is often limited to approximately 1 second(*34*). Therefore, in this study, an 8-second data window is selected to evaluate the model performance, consisting of the 4 seconds before and after the crash timestamp. Crash video data needs to be preprocessed in advance to obtain individual-level traffic dynamic data. In this study, DataFromSky is used to analyze the drone video recordings of crashes and extract the motion information of the vehicles including vehicle types, pixel positions and headings for each frame.

## 4.2. Performance of segment evolution map generation model

Following the training procedure outlined in Algorithm 1, the model training is divided into two stages. The first stage involves training a denoising U-Net, followed by the introduction of ControlNet in the second stage. During both training phases, we set the forward noise addition and reverse sampling process step $T$ to 1000 steps. The Adaptive Moment Estimation (Adam) optimizer was employed, with a batch size of 4 and an initial learning rate of 0.00005.

Additionally, to accelerate the model's convergence, all model training in this study is performed using fine-tuning techniques. Although the pre-trained Stable Diffusion models demonstrate strong image generation capabilities, they are not directly suitable for the specific task and dataset of this study as Text-to-Image or Image-to-Image models. To address this, we initialized the denoising U-Net parameters using the pre-trained U-Net model from Stable Diffusion, offering a robust starting point by leveraging the general knowledge acquired from large-scale datasets. This fine-tuning technique not only accelerates the training process, but also enhances the model's performance in the road segment map generation task.

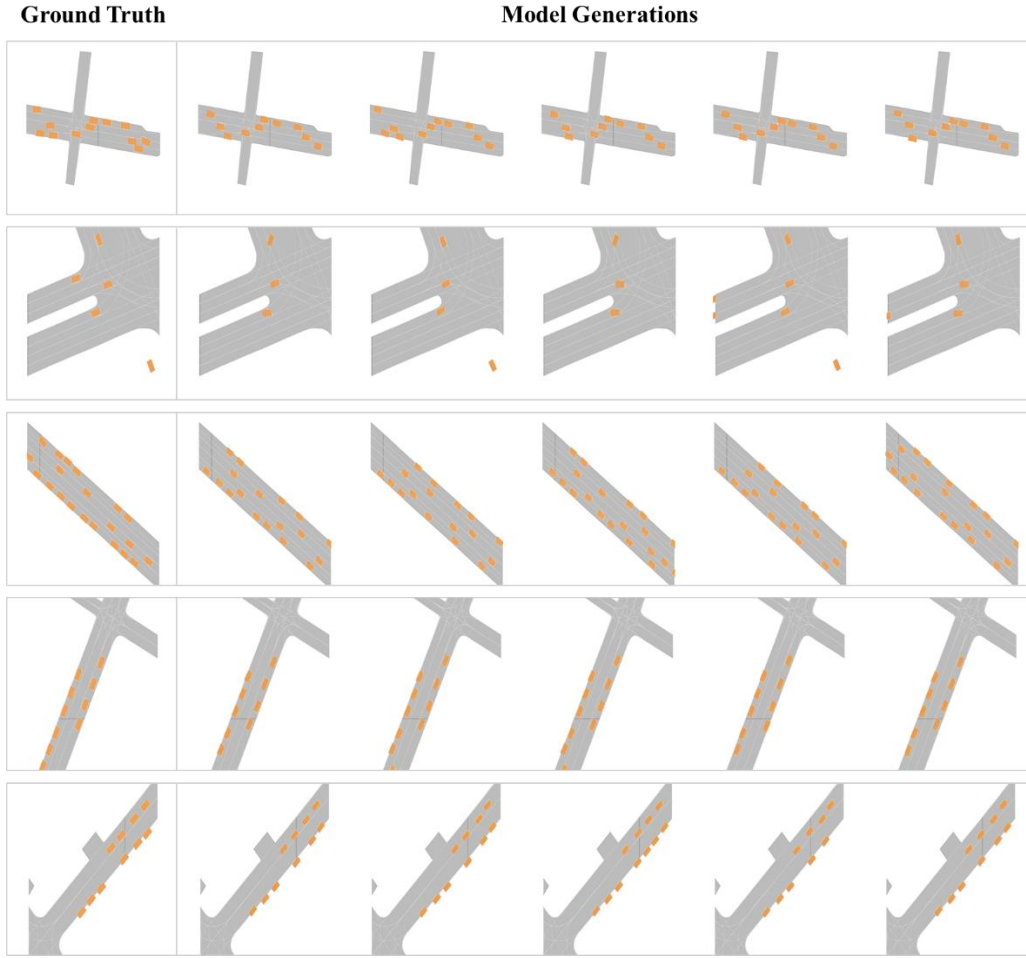

**Figure 8.** Examples of the generated segment map.

The proposed model was trained to generate plausible road segment evolution under non-crash conditions. The trained model will produce a set of reasonably evolving non-crash segment maps through multi-sampling, which serves as a reference to distinguish crash-induced deviations. Because of the highly diversified human behaviors of each driver, road segment evolution exhibits inherent diversity. Therefore, the focus of evaluating the performance of Mapfusion is on assessing whether the generated road segment maps are evolving reasonably, rather than focusing on the similarity between the generated and the ground truth. Figure 8 presents the model-generated evolution of road segment maps in several representative scenarios, through multi-sampling of diffusion model. The leftmost column displays the ground truth at the predicted time step, while the right columns present the road segment maps generated by the model with a sampling number set to 5. It is observable that the model demonstrates the capability of multi-sample generation while maintaining reasonable control over key traffic characteristics, including vehicle count, traveling direction, and lane usage.

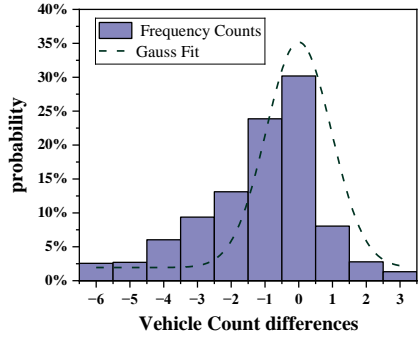

**Figure 9.** Distribution of vehicle count differences between generated and ground truth road segment maps.

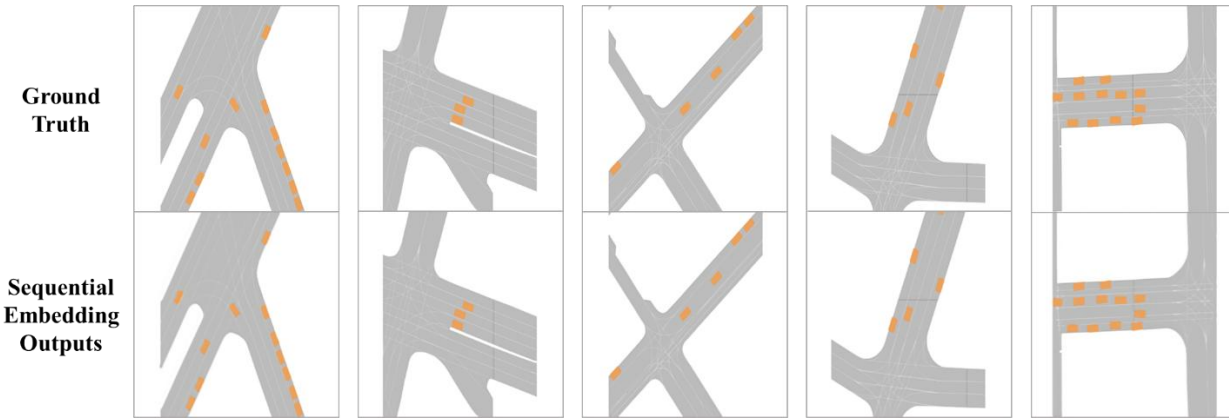

**Figure 10.** Examples of decoded sequential embedding features.

Since the number of vehicles in the generated road segment maps serves as a critical constraint, it is essential to assess the model's performance in maintaining vehicle count consistency. To achieve this, we computed the difference between the vehicle counts in the generated maps and the corresponding ground truth. A statistical analysis of these differences was conducted, as illustrated in Figure 9. The results indicate that the differences are predominantly centered around zero, further validating the effectiveness of Mapfusion in deducing the evolution of vehicles on the road segment.

### 4.3. Ablation study for ConvLSTM-based sequential embedding

To evaluate the representational capacity of the ConvLSTM-based sequential embedding components, a decoder was designed to reconstruct the feature representations extracted by the sequential embedding process from a sequence of road segment maps. This approach highlights the crucial role of the sequential embedding components in the Mapfusion for capturing the evolutionary patterns of road segment maps. Figure 10 presents examples of the decoded sequential embedding features. In each subFigure, the top row shows the ground truth at the predicted timestep. The bottom row displays the reconstructed map, generated from the learned features of the road segment maps in previous frames. From Figure 10, it can be observed that the sequential embedding can successfully identified the location, number and direction information of vehicles on the road segment map. Therefore, the input features in the cross-attention introduce accurate spatiotemporal information.

**Table 2.** Ablation studies for sequential embedding in Mapfusion.

| Model Variant | MSE | MAE |
|---|---|---|
| Baseline | 0.00046 | 0.00216 |
| Baseline w/o SE for denoising U-Net | 0.0092 | 0.0349 |
| Baseline w/ SE for ControlNet | 0.0088 | 0.0346 |

**Table 3.** Road segment generation comparison under different sampling intervals.

| Sampling interval (s) | MSE | MAE |
|---|---|---|
| 0.1 | 0.00046 | 0.00216 |
| 0.2 | 0.00048 | 0.00220 |
| 0.3 | 0.00052 | 0.00228 |
| 0.4 | 0.00056 | 0.00234 |

We further conducted ablation experiments to quantify the contribution of sequential embedding (SE) to the Mapfusion's performance. Three model variants were compared. The first variant is the original model architecture in Section 3.3, referred to as the baseline. The second corresponds removes sequential embedding from the baseline. In the third variant, the spatiotemporal information extracted by sequential embedding is incorporated into the model as the input to the ControlNet. To quantitatively evaluate model performance, we adopt Mean Squared Error (MSE) and Mean Absolute Error (MAE) as the evaluation metrics. These metrics assess pixel-level differences between the generated road segment maps and the ground truth, capturing differences in vehicle locations and counts. The numerical results in Table 2 demonstrate that sequential embedding improves generation performance. Furthermore, feeding the sequential embedding into the denoising U-Net yields better results than to ControlNet.

### 4.4. Generation performance under varying data collection frequencies

Data collection frequency varies across different sensor types and directly impacts model performance in crash detection tasks. Higher sampling frequencies have been shown to improve detection accuracy, they also increase computational demands during real-time deployment. Given that the proposed model processes temporally uniform sequences of road segment maps, it is essential to evaluate its adaptability to varying sampling frequencies. The results in Section 4.2 employ a sampling interval of 0.1 seconds. To assess the effect of temporal resolution on generation performance, the original 10 Hz data were further downsampled to four different intervals: 0.1 s, 0.2 s, 0.3 s, and 0.4 s. Separate models were trained on the datasets corresponding to each sampling frequency. For comparability, all models were evaluated by predicting the same target frame within an identical scenario. The generation results under each sampling frequency are presented in the Appendix I. The proposed generation model exhibits robust generation performance across all tested sampling frequencies. Specifically, it consistently maintains accurate inference of vehicle positions, orientations, lane alignments, and vehicle counts. Similarly, Mean Squared Error (MSE) and Mean Absolute Error (MAE) are adopted as quantitative metrics as shown in Table 3. Both MSE and MAE remain at relatively low levels across all four sampling intervals. However, the values increase gradually with larger sampling intervals, indicating that the model performs better under higher sampling frequencies.

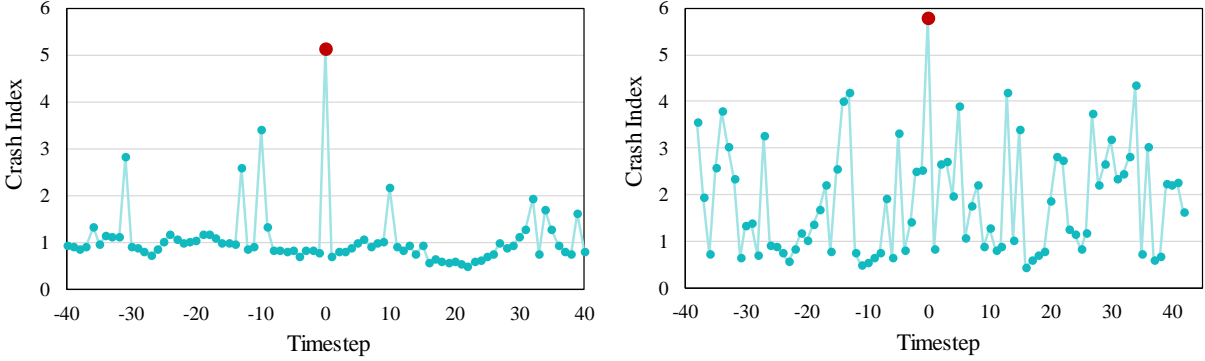

**Figure 12.** *Crash Index* distribution during crash time period.

### 4.5. Crash detection

This section evaluates the effectiveness of the proposed crash detection method in identifying real-time road segment crashes. Consistent with the generation network, an 11-frame time window was employed for crash detection. The first 10 frames of the segment maps were used as prior input for diffusion model to generate the expected road segment evolution, while the final frame served as the monitored reference to determine whether a crash had occurred. The *Crash Index* for each timestep was computed using a sliding window approach with a step size of 1. Since the exact crash occurrence times in the dataset are known in advance, we designate the crash moment as **timestep 0** to facilitate analysis. In the visual representation, crash occurrences are marked with specific symbols.

As shown in Figure 12, negative values on the time axis indicate moments preceding the crash, while positive values represent moments after the crash. *Crash Indexes* over time are presented using a connected line plot. The corresponding scatter plots demonstrate a noticeable increase in values at the timestep of the crash occurrence. This phenomenon indicates the effectiveness of our approach to detect anomalies associated with road crashes. Although the *Crash Index* exhibits fluctuations at other timesteps, the peak values of these variations remain lower than those observed at actual crash moments. Moreover, these fluctuations align with real-world traffic dynamics. Based on these findings, it can be concluded that the proposed framework is capable of effectively performing crash detection.

It is worth noting that the value ranges of *Crash Index* for these two crash samples on the Hurong highway were similar during both the crash and non-crash periods. Although determining a judgment threshold with a high level of confidence remains challenging due to the limited availability of crash data, the fluctuation patterns of the two curves still provide valuable insights into crash occurrences. Therefore, even without prior knowledge of an exact *Crash Index* threshold, the proposed crash detection framework demonstrates a relatively effective crash detection capability.

### 4.6. Discussion

The performance of the proposed framework can be compared with existing trajectory-based and macroscopic approaches. Compared to trajectory-based methods, our approach does not rely on continuous vehicle identity tracking. This makes it more robust in cases of tracking failure or vehicle ID switching, which often degrade the performance of trajectory-based models. Furthermore, trajectory-based models, whether modeling pairwise interactions or individual trajectories, become computationally expensive as vehicle density increases. In contrast, our method maintains consistent efficiency by operating on aggregated road segment representations. Besides, existing modeling approaches for crash detection, including both machine learning and deep learning approaches, require the extraction of macro- or micro-level traffic features from both before and after the crash. In scenarios with extremely limited crash data, such methods become difficult or even infeasible to train effectively. With only two recorded crash sequences, conventional classifiers such as SVM and XGBoost failed to train stable models, highlighting the limitations of supervised approaches under data scarcity. However, our generative framework, by leveraging non-crash data for learning motion patterns, remains robust even in low-label environments.

As for traditional methods based on the numerical fluctuations of macroscopic variables, they offer simplicity and low computational costs, but lack granularity and cannot localize specific vehicles involved in anomalous behavior. By contrast, our method not only identifies the time of a potential crash but also can provide the corresponding segment map. This visual representation allows for intuitive inspection of abnormal vehicle behavior, thus supporting both algorithmic detection and human-in-the-loop verification.

## 5. Conclusions

This paper proposes an explicit two-stage trajectory-free crash detection method, utilizing individual-level traffic dynamic data based on a novel diffusion model framework. After encoding vehicle data into road segment maps, Mapfusion applies forward diffusion and reverse denoising to generate segment evolutions under non-crash conditions in the first stage. The second stage achieved crash detection by quantitatively comparing the monitored road segment map with the generated non-crash maps from the first stage. Experimental results demonstrate that the proposed method effectively identifies crash occurrences and reveal the correlation between *Crash Index* distribution and crash detection.

Furthermore, the proposed road segment map representation addresses the challenges of trajectory acquisition while preserving multi-participant information. Additionally, the crash detection approach leverages extrapolation from normal driving scenarios. This characteristic enables it to fully capture the wide range of possible crash scenarios and effectively identify various types of crash events deviating from non-crash driving patterns, including both single-vehicle and multi-vehicle crashes. As a result, this method mitigates the limitations stemming from the scarcity of available crash samples and the absence of prior knowledge regarding diverse crash categories. The generative two-stage crash detection framework can effectively address the challenges of large labeling costs, class imbalance, and the lack of comprehensiveness in the training set for anomalous samples.

This study has several limitations that warrant further investigation. The proposed Mapfusion processes uniformly sampled data as input and does not account for scenarios involving non-uniform sampling intervals or frame loss during traffic dynamics data collection. Future work will focus on enhancing the sequential embedding components to handle irregularly sampled inputs. Additionally, the present framework is designed primarily for discrete road segments. A multi-agent model that accounts for contiguous road segments will be constructed in the future to facilitate a cooperative detection of road crashes. Finally, given the limited crash dataset used in this study, the results should be interpreted as a proof-of-concept validation, and further validation with larger-scale data is planned for future work.

## Declaration of competing interest

The authors declare that they have no known competing financial interests or personal relationships that could have appeared to influence the work reported in this paper.

## Acknowledgment

This work was jointly supported by the projects of the National Natural Science Foundation of China [grant number: 52232012], the Natural Science Foundation of the Jiangsu Province [grant number: BK20231428], the Fundamental Research Funds for the Central Universities [grant number: 2242022R40024], and Nanjing Municipal Program for Technological Innovation by Overseas Scholar [grant number: 1121002326].

## Author contributions

**Weiying Shen:** Writing – original draft, Validation, Methodology, Investigation, Conceptualization. **Hao Yu:** Writing – review & editing, Validation, Supervision, Methodology, Funding acquisition, Conceptualization. **Yu Dong:** Writing – original draft, Validation, Methodology, Investigation. **Pan Liu:** Writing – review & editing, Supervision, Funding acquisition, Conceptualization. **Yu Han:** Writing – review & editing, Data curation. **Xin Wen:** Writing – review & editing, Data curation.

# Appendix I

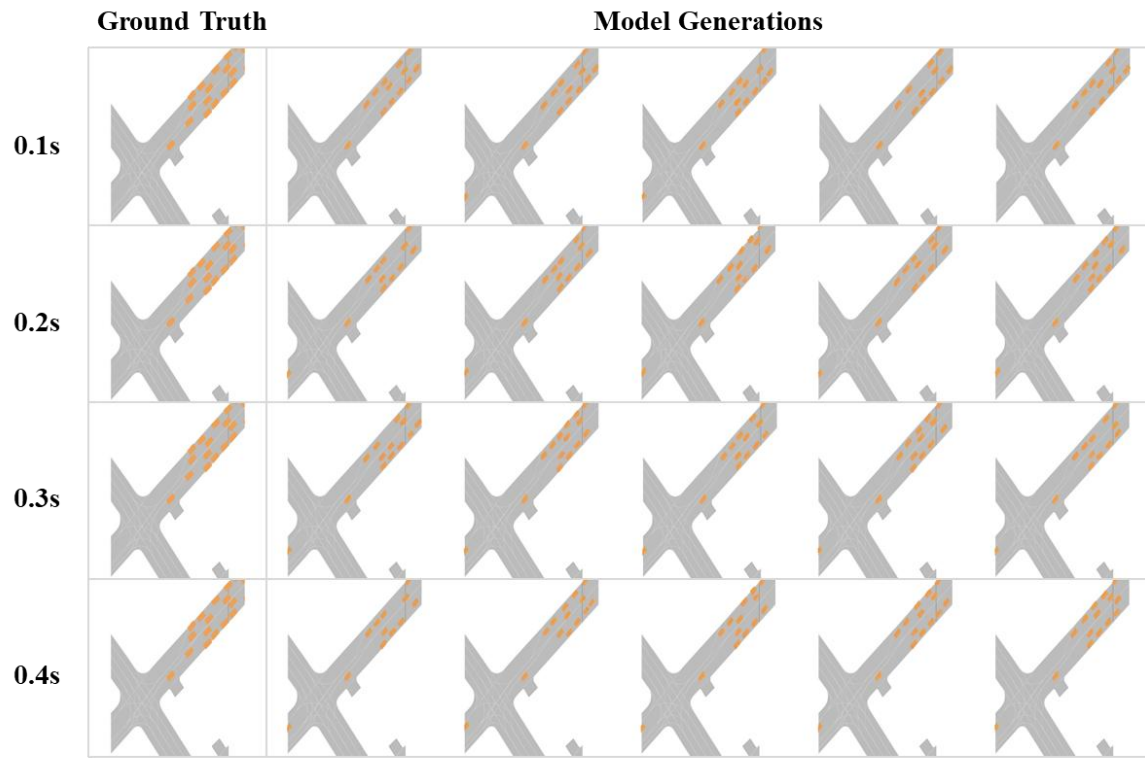

(a) Scenario 1

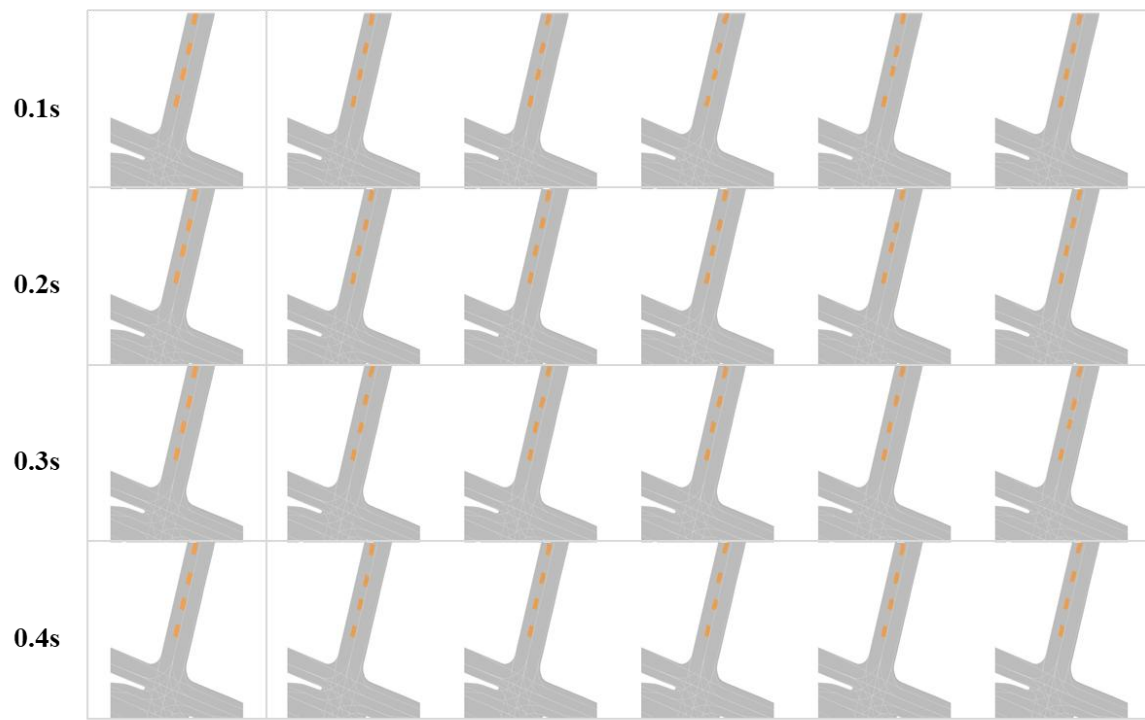

(b) Scenario 2

# References

1. Payne, H., and S. Tignor. Freeway Incident Detection Algorithms Based on Decision Trees with States. *Transportation Research Record: Journal of the Transportation Research Board*, 1978.
2. Kalair, K., and C. Connaughton. Anomaly Detection and Classification in Traffic Flow Data from Fluctuations in the Flow–Density Relationship. *Transportation Research Part C: Emerging Technologies*, 2021. 127: 103178.
3. Chen, T. -Y., and R. -C. Jou. Using HLM to Investigate the Relationship between Traffic Accident Risk of Private Vehicles and Public Transportation. *Transportation Research Part A: Policy and Practice*, 2019. 119: 148–161.
4. Moriano, P., A. Berres, H. Xu, and J. Sanyal. Spatiotemporal Features of Traffic Help Reduce Automatic Accident Detection Time. *Expert Systems with Applications*, 2024. 244: 122813.
5. Dogru, N., and A. Subasi. Traffic Accident Detection Using Random Forest Classifier. Presented at *2018 15th Learning and Technology Conference (L&T)*, Jeddah, Saudi Arabia, 2018.
6. Liyanage, Y. W., D. -S. Zois, and C. Chelmis. Robust Freeway Accident Detection: A Two–Stage Approach. Presented at *ICASSP 2019-2019 IEEE International Conference on Acoustics, Speech and Signal Processing (ICASSP)*, Brighton, UK, 2019.
7. Yang, D., Y. Wu, F. Sun, J. Chen, D. Zhai, and C. Fu. Freeway Accident Detection and Classification Based on the Multi-Vehicle Trajectory Data and Deep Learning Model. *Transportation Research Part C: Emerging Technologies*, 2021. 130: 103303.
8. Wang, L., F. Zhou, Z. Li, W. Zuo, and H. Tan. Abnormal Event Detection in Videos Using Hybrid Spatio-Temporal Autoencoder. Presented at *25th IEEE International Conference on Image Processing (ICIP)*, Athens, Greece, 2018.
9. Chakraborty, P., A. Sharma, and C. Hegde. Freeway Traffic Incident Detection from Cameras: A Semi-Supervised Learning Approach. Presented at *21st International Conference on Intelligent Transportation Systems (ITSC)*, Maui, HI, USA, 2018.
10. Chen, X., H. Xu, M. Ruan, M. Bian, Q. Chen, and Y. Huang. SO-TAD: A Surveillance-Oriented Benchmark for Traffic Accident Detection. *Neurocomputing*, 2025. 618: 129061.
11. Fang, J., J. Qiao, J. Xue, and Z. Li. Vision-Based Traffic Accident Detection and Anticipation: A Survey. *IEEE Transactions on Circuits and Systems for Video Technology*, 2024. 34(4): 1983–1999.
12. Xia, Y., N. Qian, L. Guo, and Z. Cai. CF-SOLT: Real-Time and Accurate Traffic Accident Detection Using Correlation Filter-Based Tracking. *Image and Vision Computing*, 2024. 152: 105336.
13. Papadopoulos, A., A. Sersemis, G. Spanos, A. Lalas, C. Liaskos, K. Votis, and D. Tzovaras. Lightweight Accident Detection Model for Autonomous Fleets Based on GPS Data. *Transportation Research Procedia*, 2024. 78: 16–23.
14. Wu, J., Y. Zhang, and H. Xu. A Novel Skateboarder-Related near-Crash Identification Method with Roadside LiDAR Data. *Accident Analysis & Prevention*, 2020. 137: 105438.
15. Afizi, M., M. Shukran, M. Naim, M. Mohd, M. Nazri, M. Khairuddin, and K. Maskat. Developing a Framework for Accident Detecting and Sending Alert Message Using Android Application. *International Journal of Engineering and Technology (UAE)*, 2018. 7: 66–68.
16. Zhu, M., H. Yang, C. Liu, Z. Pu, and Y. Wang. Real-Time Crash Identification Using Connected Electric Vehicle Operation Data. *Accident Analysis & Prevention*, 2022. 173: 106708.
17. Huang, Y., W. Wei, H. Yang, Q. Wu, and K. Xu. Intelligent Algorithms for Incident Detection and Management in Smart Transportation Systems. *Computers and Electrical Engineering*, 2023. 110: 108839.

18. Ma, Y., J. Zhang, J. Lu, S. Chen, G. Xing, and R. Feng. Prediction and Analysis of Likelihood of Freeway Crash Occurrence Considering Risky Driving Behavior. *Accident Analysis & Prevention*, 2023. 192: 107244.
19. Zhang, T., and P. J. Jin. A Longitudinal Scanline Based Vehicle Trajectory Reconstruction Method for High-Angle Traffic Video. *Transportation Research Part C: Emerging Technologies*, 2019. 103: 104–128.
20. Li, P., X. Pei, Z. Chen, X. Zhou, and J. Xu. Human-like Motion Planning of Autonomous Vehicle Based on Probabilistic Trajectory Prediction. *Applied Soft Computing*, 2022. 118: 108499.
21. Lee, N., W. Choi, P. Vernaza, C. B. Choy, P. H. S. Torr, and M. Chandraker. DESIRE: Distant Future Prediction in Dynamic Scenes with Interacting Agents. Presented at *2017 IEEE Conference on Computer Vision and Pattern Recognition (CVPR)*, Honolulu, HI, USA, 2017.
22. Goodfellow, I., J. Pouget-Abadie, M. Mirza, B. Xu, D. Warde-Farley, S. Ozair, A. Courville, and Y. Bengio. Generative Adversarial Nets. *NIPS'14: Proceedings of the 28th International Conference on Neural Information Processing Systems*, 2014. 2: 2672–2680.
23. Li, C., G. Feng, Y. Li, R. Liu, Q. Miao, and L. Chang. DiffTAD: Denoising Diffusion Probabilistic Models for Vehicle Trajectory Anomaly Detection. *Knowledge-Based Systems*, 2024. 286: 111387.
24. Shi, H., S. Dong, Y. Wu, Q. Nie, Y. Zhou, and B. Ran. Generative Adversarial Network for Car Following Trajectory Generation and Anomaly Detection. *Journal of Intelligent Transportation Systems*, 2024. 1–14.
25. Ho, J., A. Jain, and P. Abbeel. Denoising Diffusion Probabilistic Models. *ArXiv*, 2020. 2006.11239.
26. Rombach, R., A. Blattmann, D. Lorenz, P. Esser, and B. Ommer. High-Resolution Image Synthesis with Latent Diffusion Models. *ArXiv*, 2021. 2112.10752.
27. Ramesh, A., P. Dhariwal, A. Nichol, C. Chu, and M. Chen. Hierarchical Text-Conditional Image Generation with CLIP Latents. *ArXiv*, 2022. 2204.06125.
28. Ho, J., T. Salimans, A. Gritsenko, W. Chan, M. Norouzi, and D. J. Fleet. Video Diffusion Models. *ArXiv*, 2022. 2204.03458.
29. Zhang, L., A. Rao, and M. Agrawala. Adding Conditional Control to Text-to-Image Diffusion Models. *ArXiv*, 2023. 2302.05543.
30. Shi, X., Z. Chen, H. Wang, D. Y. Yeung, W. C. Wong, and W. Woo. Convolutional LSTM Network: A Machine Learning Approach for Precipitation Nowcasting. *ArXiv*, 2015. 1506.04214.
31. Heusel, M., H. Ramsauer, T. Unterthiner, B. Nessler, and S. Hochreiter. GANs Trained by a Two Time-Scale Update Rule Converge to a Local Nash Equilibrium. *NIPS'17: Proceedings of the 31st International Conference on Neural Information Processing Systems*, 2017. 6629–6640.
32. Dosovitskiy, A., L. Beyer, A. Kolesnikov, D. Weissenborn, X. Zhai, T. Unterthiner, M. Dehghani, M. Minderer, G. Heigold, S. Gelly, J. Uszkoreit, and N. Houlsby. An Image Is Worth 16x16 Words: Transformers for Image Recognition at Scale. *ArXiv: Computer Vision and Pattern Recognition*, 2020. 2010.11929.
33. Wilson, B., W. Qi, T. Agarwal, J. Lambert, J. Singh, S. Khandelwal, B. Pan, R. Kumar, A. Hartnett, J. Pontes, D. Ramanan, P. Carr, and J. Hays. Argoverse 2: Next Generation Datasets for Self-Driving Perception and Forecasting. *Proceedings of the Neural Information Processing Systems Track on Datasets and Benchmarks (NeurIPS Datasets and Benchmarks 2021)*, 2021.
34. Wang, Y., C. Xu, P. Liu, Z. Li, and K. Chen. Assessing the Predictability of Surrogate Safety Measures as Crash Precursors Based on Vehicle Trajectory Data Prior to Crashes. *Accident Analysis & Prevention*, 2024. 201: 107573.